Deep-learning models improve on community-level diagnosis for common congenital heart disease lesions

Rima Arnaout MD[1], Lara Curran MBBS BSc[1], Erin Chinn MS[1], Yili Zhao PhD RDCS[1], and Anita Moon-Grady MD[1]

[1]University of California, San Francisco, San Francisco, California, USA

**Abstract**

Prenatal diagnosis of tetralogy of Fallot (TOF) and hypoplastic left heart syndrome (HLHS), two serious congenital heart defects, improves outcomes and can in some cases facilitate *in utero* interventions[1,2]. In practice, however, the fetal diagnosis rate for these lesions is only 30-50 percent in community settings[3-6]. Improving fetal diagnosis of congenital heart disease is therefore critical. Deep learning is a cutting-edge machine learning technique for finding patterns in images but has not yet been applied to prenatal diagnosis of congenital heart disease. Using 685 retrospectively collected echocardiograms from fetuses 18-24 weeks of gestational age from 2000-2018, we trained convolutional and fully-convolutional deep learning models in a supervised manner to (*i*) identify the five canonical screening views of the fetal heart and (*ii*) segment cardiac structures to calculate fetal cardiac biometrics. We then trained models to distinguish by view between normal hearts, TOF, and HLHS. In a holdout test set of images, F-score for identification of the five most important fetal cardiac views was 0.95. Binary classification of unannotated cardiac views of normal heart vs. TOF reached an overall sensitivity of 75% and a specificity of 76%, while normal vs. HLHS reached a sensitivity of 100% and specificity of 90%, both well above average diagnostic rates for these lesions[3,6]. Furthermore, segmentation-based measurements for cardiothoracic ratio (CTR), cardiac axis (CA), and ventricular fractional area change (FAC) were compatible with clinically measured metrics for normal, TOF, and HLHS hearts. Thus, using guideline-recommended imaging, deep learning models can significantly improve detection of fetal congenital heart disease compared to the common standard of care.

**Introduction**

Congenital heart disease (CHD), is both the most common birth defect and yet rare overall, affecting one percent of live births[3]. CHD can be asymptomatic in fetal life but cause significant morbidity and mortality after birth[3,7,8]. The earlier CHD is diagnosed, the better the outcomes and therapeutic options at birth[9-11]. There are also increasingly available and effective *in utero* therapies for specific CHD lesions (e.g. in *utero aortic* valvuloplasty for HLHS) which can significantly improve the natural history of disease[1,2]. These potential benefits all rely on accurate fetal diagnosis of CHD.

Fetal screening ultrasound is recommended for every pregnant woman worldwide[12,13] between 18-24 weeks of gestation and provides five clinically recommended screening views of the heart (Figure 1) that in theory can diagnose over 90 percent of complex congenital heart disease[14]. In practice, however, the fetal diagnosis rate for congenital heart disease in the community is 30-50 percent [3-5], even where fetal ultrasound is universal[5,12,13].

We hypothesized that the main reason for this startling diagnosis gap is inadequate/uneven expertise in interpreting fetal cardiac images, due to the diagnostic challenge presented by a small and fast-beating fetal heart and due to relatively low exposure to congenital heart disease among caregivers (owing to its low prevalence). Small, single-center clinical quality control programs can increase detection of CHD up to 100 percent[15,16], but such programs are difficult to sustain and scale. We therefore decided to test whether deep-learning image analysis could improve upon diagnosis rates commonly encountered in community practice, even when trained only on a relatively small number of clinically relevant imaging studies.

Deep learning is good at pattern recognition in images[17,18] and has been applied to adult cardiac ultrasound, besting clinicians task-for-task on view classification using small, downsampled datasets[19]. However, deep learning has not yet been widely applied to CHD or fetal ultrasound, nor in general to clinical conditions that are by definition rare, no matter how widely the net is cast for training data. We hypothesized that using input data curated according to clinical guidelines (i.e., selecting only the five screening cardiac views) would allow our models to detect diagnostic signals on small datasets.

Here, we demonstrate a pipeline for image processing and deep-learning assisted view identification, quantification, and view-based diagnosis in order to improve fetal CHD diagnosis for the most common CHD lesion (TOF) and the lesion currently most amenable to in utero intervention (HLHS).

**Methods**

Dataset. All data were obtained and de-identified, with waived consent in compliance with the Institutional Review Board (IRB) at the University of California, San Francisco (UCSF). 685 fetal echocardiograms performed between 2000 and 2018 were extracted from UCSF's database; 493 were of structurally normal hearts, 87 were of TOF, and 105 were of HLHS. Images came from GE, Philips, and Siemens ultrasound machines. Inclusion criteria were singleton fetuses of 18-24 weeks of gestational age. Presence of significant non-cardiac malformations (e.g. congenital diaphragmatic hernia) were excluded. Echocardiogram pathology was determined both by review of the clinical report, as well as expert visual review of each echocardiogram. Normal fetal echocardiograms were negative for structural heart disease, fetal arrhythmia, maternal diabetes, presence or history of abnormal nuchal translucency measurement, non-cardiac congenital malformations, and postnatal CHD diagnosis.

Data Processing. DICOM-formatted image were deidentified as previously described[19]. Axial sweeps of the thorax from fetal echocardiogram were split into constituent frames at 300 by 400 pixel resolution and labeled by view—3-vessel trachea (3VT), 3-vessel view (3VV), apical 5-chamber (A5C), apical 4-chamber (A4C), and abdomen (ABDO) (Figure 1)—resulting in 29,650 images for view classification training. In order to ensure that the images chosen were of sufficient diagnostic quality without disqualifying adequate images, we employed the judgement of our clinical expert. Images were cropped to 180 by 240 pixels, and then downsampled to 60 by 80 pixels, normalized with respect to greyscale value, and split into training and holdout test datasets in approximately an 80:20 ratio. Each dataset contained images from separate echocardiographic studies, to maintain sample independence. The number of images in training and test datasets for view classification were 23,350 and 6,300 images, respectively. For lesion detection, images were split into approximately 80:20 or 90:10 ratios for training and holdout test sets. 25,726 and 5,423 images were used to train and test normal vs. TOF ; 43,667 and 4,829 images were used to train and test normal vs. HLHS; and 59,487 and 7,202 images were used to train and test normal vs. either lesion. For training fetal structural and functional measurements, custom software using Python's OpenCV library (https://opencv.org/) was written to label thorax, heart, right atrium, right ventricle, left atrium, left ventricle, spine, and interventricular septum from A4C images. For models classifying CHD, images from normal hearts were trained against the CHD lesions mentioned above. Coding was done in Python.

Model Architecture and Training. *View classification.* View classification was performed as previously described[19], with the following modifications to data augmentation: rotation range 10, width/height shift range 0.3, horizontal and vertical flipping allowed, shear of 0.01, and zoom of 0.05. Models were trained for 150-200 epochs. Trainings were performed on Amazon's EC2 platform with a GPU instance g3.4xlarge and took 12-18h. Roughly equal proportions of data classes were used in training datasets. For view classification, a training dataset in which view labels were randomized was used as a negative control, resulting in an F-score of 0.19 (commensurate with random chance among five classes).

*Quantification of cardiothoracic ratio, chamber fractional area change, and cardiac axis.* A4C images with labeled cardiothoracic structures as above were used as training inputs to a U-Net[20] neural network architecture. These structures were predicted by U-Net, and predictions were used to calculate structural/functional metrics as follows: cardiothoracic ratio was measured as the ratio of the heart perimeter to the thorax perimeter. Fractional area change for each cardiac chamber was calculated as [maximum area – minimum area]/[maximum area]. Cardiac axis was calculated as the angle between a line centered on the spine and a line centered on the interventricular septum. Images with grossly mis-labeled structures (e.g. heart outside the thorax) were excluded from measurement analysis.

*Diagnosis of CHD lesions.* For each of the five cardiac views of interest, binary or multi-classifiers were trained and tested between normal hearts and TOF; normal and HLHS; and normal vs TOF and HLHS together. For a given ultrasound study, this resulted in five different predictions for a CHD lesion, one for each view (if multiple images were available per view in a given study, confidence probabilities among those images were averaged). For a given study, these predictions can be represented as a five-dimensional bitstring. Clinically, the ideal normal study is represented by 00000 (all views normal), while TOF and HLHS are either 11100 or 11110 (abnormality in each of the first three or four views). To arrive at a composite diagnostic score, we selected views with C-statistics > 60 to create a 4-dimensional vector. We embedded vectors from test ultrasound images in Euclidean space and took the Manhattan distance from the origin (all views normal) as the composite score, defining the score threshold at ≥2 based on the clinically ideal vectors above.

Model Evaluation. Overall accuracy, per-class accuracy, average accuracy, confusion matrices, F-scores, receiver operator characteristics, C-statistics, were calculated as previously described[19]. For performance analysis of segmentation models, Jaccard indices were calculated according to the standard definition (the intersection of predicted and labeled structures divided by their union). Sensitivity, specificity, accuracy, positive predictive value, and negative predictive value of binary diagnostic tasks were calculated from composite scores. Concordance of predicted quantitative measurements were compared to clinical and/or ground truth measures using the Mann-Whitney U test.

Data Availability. UCSF policy precludes sharing of this patient data, due to its sensitive nature.

**Results**

To test whether deep-learning models can help improve fetal CHD detection from its current rate of 30-50 percent[3,4], we created a deep-learning pipeline to (i) identify the five recommended views of the heart, (ii) use these views to calculate cardiothoracic ratio, cardiac axis, and fractional area change for each cardiac chamber, and (iii) provide classification of normal heart vs. TOF and/or HLHS CHD lesions.

Identification of the five views of the heart used in fetal CHD screening is a prerequisite for disease classification. We trained a model to pick the five screening views from fetal surveys with and F-score of 0.95 (Figure 2). In contrast to a previous view classifier trained for adult echocardiograms[19], the performance on this view classifier was achieved on individual image frames. Using a U-Net[20] neural network architecture, we trained a model to detect cardiothoracic structures from A4C images of the heart (Figure 3). These structures were used to calculate cardiothoracic ratio (CTR), cardiac axis (CA), and fractional area change (FAC) for each cardiac chamber. Examples of segmented structures are in Figure 3. Predictably, Jaccard similarities were higher for more highly represented pixel classes.

*Cardiothoracic ratio (CTR) in normal hearts.* Per-class Jaccard similarities were 0.81, 0.76, and 0.96 respectively for cardiac, thoracic, and background pixels. Normal cardiothoracic ratios reported in the literature range from 0.5 to 0.6[3]. Studies in our pilot dataset had CTRs measured clinically of 0.52±0.04. Mann-Whitney U (MWU) testing showed no statistical differences among clinically measured CTR and labeled CTR in our dataset (p-value 0.31). The mean predicted CTR for our data was 0.53±0.05, similar to the clinically measured and labeled CTRs (MWU p-value 0.23 comparing clinical and predicted; MWU p-value 0.11 comparing labeled and predicted).

*Cardiac axis (CA) in normal hearts.* In our test set, per-class Jaccard similarities were 0.72, 0.44, 0.63, 0.72, and 0.94 for heart, interventricular septum, spine, thoracic, and background pixels, respectively. A normal cardiac axis is 45 degrees (46±9 in published studies[21]), but in clinical practice, measurements ranging from 25-70 are accepted as normal[3]. It is not clear whether this wider range is due to physiologic variation alone or measurement error in clinical practice. In our pilot dataset, the average labeled CA was 46±5 degrees. Mann-Whitney U (MWU) testing showed no differences among clinically measured CA and labeled CA in our dataset (p-value 0.4). Mean predicted CA was 44±11 degrees, similar to our labels (MWU p-value 0.4).

*Chamber fractional area change (FAC) in normal hearts.* In addition to the five still-image views, it is best practice to also obtain a video of the A4C view in order to assess cardiac function[3]. FAC quantifies what is typically a qualitative assessment when performed at all in fetal surveys. Jaccard similarities for cardiac chambers were 0.66, 0.61, 0.73, 0.8, and 0.98 for left ventricle, right ventricle, left atrium, right atrium, and background, respectively. From a study measuring 70 normal fetuses, left and right ventricular FAC for fetuses of 18-24 weeks gestation averaged 0.34±0.01 and 0.33±0.02, respectively[22]. Data for fetal atrial FAC were not found in the literature. In our labeled dataset, FAC were 0.53±0.08 for left ventricle, 0.43±0.11 for right ventricle, 0.53±0.05 for left atrium, and 0.46±0.08 for right atrium. Model-predicted chamber FAC were similar to labels, with MWU p-values of 0.15, 0.41, 0.19, and 0.29 for left ventricle, right ventricle, left atrium, and right atrium, respectively.

*Classification of normal, TOF, and HLHS hearts.* We used composite score across binary classifications of views to distinguish (i) between normal and TOF hearts, with a sensitivity and specificity of 75% and 76%; (ii) between normal and HLHS hearts, with a sensitivity and specificity of 100% and 90%; and (iii) between normal and either TOF or HLHS hearts, with a sensitivity and specificity of 89% and 83%.

We sought to evaluate whether our models' decision making correlated with clinical insights. Plotting ROC curves for diagnosis by view, we observed that for TOF, C-statistics were highest for the three views from which TOF is always clinically appreciable: those of 3VT (0.80), 3VV (0.89), and A5C (0.84) (Figure 4a). A4C can either be normal or abnormal (increased CA) and had a corresponding C-statistic of 0.69. while abdomen, which is never abnormal in TOF, had a C-statistic of 0.51. Similarly, C-statistics for HLHS imitated clinical diagnosis, ranging from 0.88 to 0.94 for the four views from which this lesion can be diagnosed, while for abdomen it was 0.4 (Figure 4b). Therefore, the model relied on the same views for specific CHD lesion diagnosis as human experts would. For estimation of normal vs either TOF or HLHS, C-statistics across the views were 0.91, 0.94, 0.85, 0.85, and 0.55 for 3VT, 3VV, A5C, A4C, and abdomen, respectively (Figure 4c).

Using our segmentation models, we predicted measurements from TOF and HLHS A4C views. Mean CA for TOF was 58±10 degrees (range 42-76) (Figure 4e), while mean CTR was 0.52±0.07 (range 0.3-0.67). Consistent with the literature[23], CTR for TOF was similar to CTR for normal hearts (MWU p-value 0.46), while CA was different from normal (MWU p-value 7.4e-6).

**Discussion**

In an era of unprecedented ability to treat CHD earlier and earlier and clear benefit to early diagnosis and treatment, the need for accurate, scalable fetal screening for CHD has never been stronger[24]. To address the need for improved CHD diagnosis, we used guidelines-based fetal imaging to train an ensemble of deep learning models to identify and quantify the fetal heart. On three diagnostic tasks, our model beat accuracies for common CHD detection reported in the literature and approached the performance of national experts in fetal cardiology. This increase in performance compared to commonly reported detection rates[3,6] using only recommended screening images supports the hypothesis that current recommendations for imaging are sufficient, but that expert interpretation of these images is difficult to achieve and maintain at worldwide scale.

While it is the most common birth defect, complex CHD is still relatively rare, affecting only about one percent of neonates. Deep learning is a data-hungry machine learning method, but we showed that deep learning on images can be used on rare diseases to significantly boost diagnosis from what is commonly found in practice, using a surprisingly small number of ultrasound studies. We did this by choosing our input data according to clinical recommendations for only five cardiac views of interest rather than the entire ultrasound. This strategy allowed us to reduce the size of the input data to our diagnostic model and thereby achieve computational efficiency in training and in subsequent predictions on new data. This efficiency in prediction is key to translating this work toward real-world and resource-poor settings. Second, our use of clinical guidelines to define our image inputs can help ease adoption of deep-learning assisted diagnosis among providers. While quantitative measures of fetal structure and function approximated clinical metrics and followed patterns found in normal and diseased hearts, further validation of these measures and their underlying segmentation measures is needed. Undoubtedly, performance of our model could be improved by training on more studies, and with further improvements in neural network architecture.

**Conclusions**

We showed that an ensemble of machine learning models trained on guideline-recommended imaging can significantly improve detection of fetal congenital heart disease—a rare disease—compared to the common standard of care. We demonstrated that the current diagnosis gap for fetal CHD is largely fueled by problems applying accurate and reproducible interpretation to images. We look forward to testing and refining these models in larger populations in an effort to democratize the expertise of fetal cardiology experts to providers and patients worldwide.

**Acknowledgements:** We thank Dr. Michael Brook for help with database access.

**Funding:** R.A. was funded by AHA 17IGMV33870001, NIH K08HL125945, and a UCSF Catalyst award.

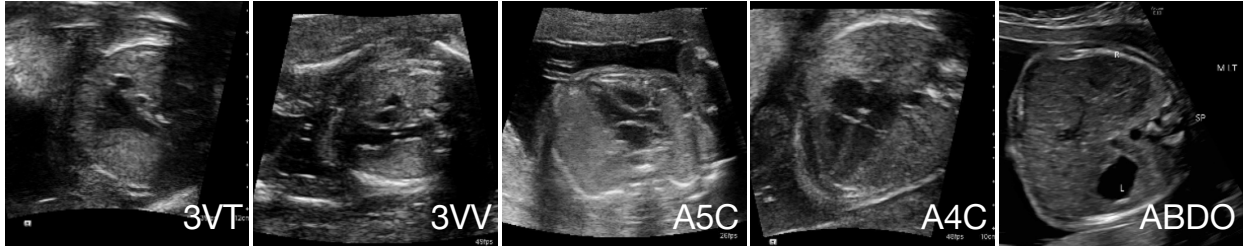

Figure 1. The five canonical screening views of the fetal heart. 3VT, 3-vessel trachea view; 3VV, 3-vessel view; A5C apical 5-chamber view; A4C apical 4-chamber view; ABDO abdomen view.

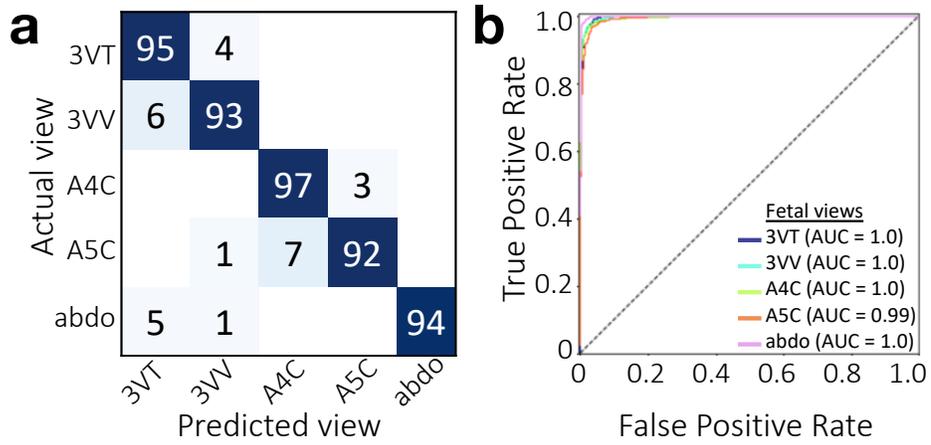

Figure 2. Confusion matrix (a) and AUC (b) for fetal 5-view classification.

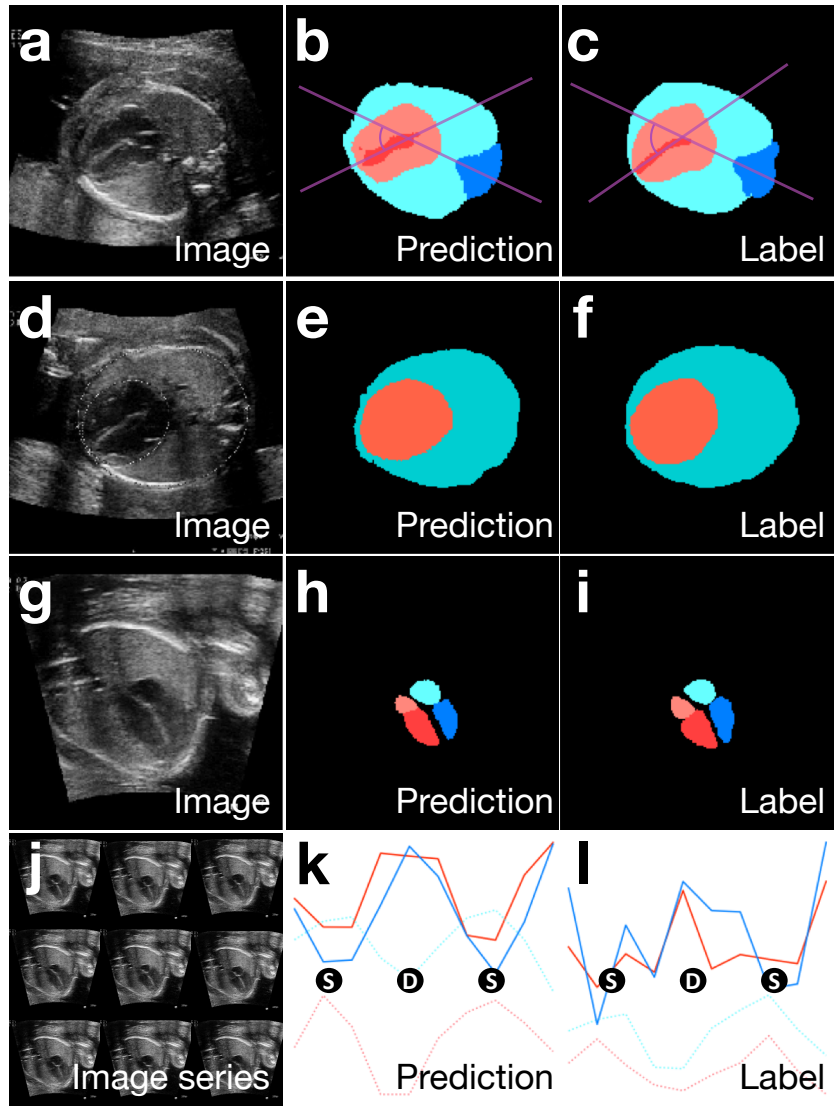

Figure 3. Fetal cardiac structure and function measurements using segmentation models. Example native image, prediction, and ground truth label: (a)-(c) for cardiac axis. Teal: thorax, blue: spine, pink: heart, red: septum. (d)-(f) for cardiothoracic ratio. Teal: thorax, red: heart. (g)-(i) for cardiac chambers. Teal: RA, blue: RV, pink: LA, red: LV. Chamber segmentation for a sequence of images in a video (j) allows plots of chamber area over time (k)-(l) and identification of image frames in ventricular systole (S) and diastole (D). (k) Predicted cardiac cycle. (l) Cardiac cycle from ground truth labeling.

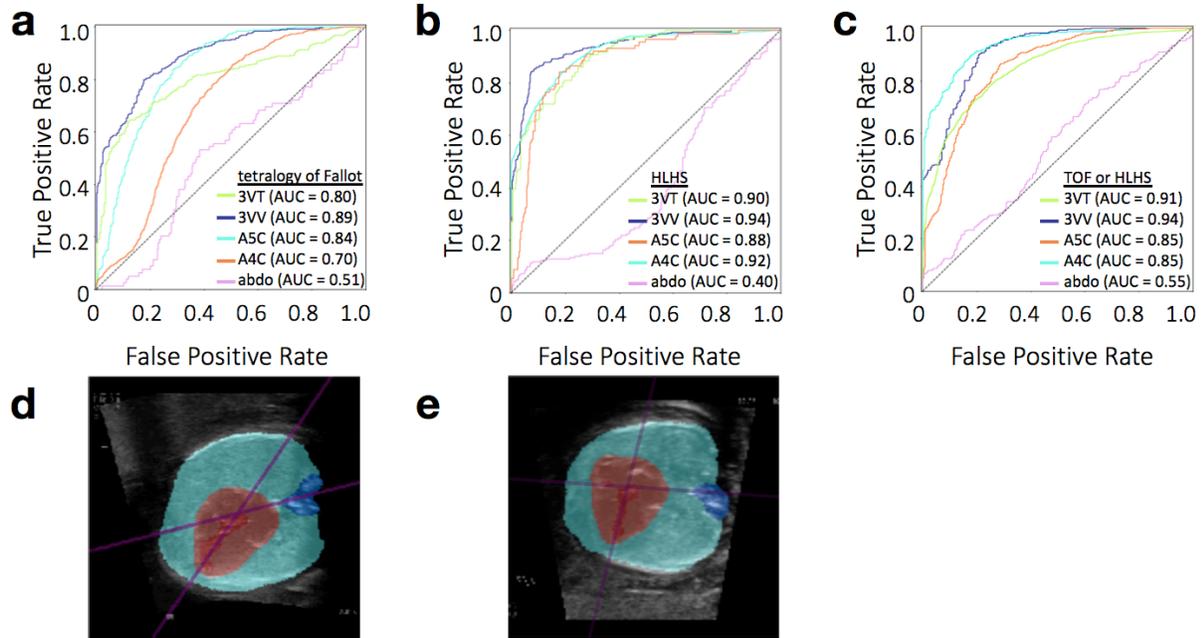

Figure 4. Classification performance for diagnosis of TOF and HLHS. AUC for TOF (a), HLHS (b), and TOF or HLHS (c), by view. A representative normal heart (d) has a smaller cardiac axis than a representative TOF heart (e) in the A4C view.